\documentclass[10pt]{article}
\usepackage{amsmath, setspace, geometry}
\usepackage{graphicx}
\usepackage{lineno}
\usepackage{natbib}
\usepackage{url}

\begin{document}

\noindent
{\LARGE\textbf{Classification of the lunar surface pattern by AI architectures: Does AI see a rabbit in the Moon?}} 
\\
\\
{\Large Daigo Shoji$^{1}$ (shohji.daigo@jaxa.jp)}\\
1. Institute of Space and Astronautical Science (ISAS), Japan Aerospace Exploration Agency (JAXA), 3-1-1, Yoshinodai, Chuo-ku, Sagamihara, Kanagawa, Japan.\\
\doublespacing

\section*{Abstract}
In Asian countries, there is a tradition that a rabbit, known as the Moon rabbit, lives on the Moon. Typically, two reasons are mentioned for the origin of this tradition. The first reason is that the color pattern of the lunar surface resembles the shape of a rabbit. The second reason is that both the Moon and rabbits are symbols of fertility, as the Moon appears and disappears (i.e., waxing and waning) cyclically and rabbits are known for their high fertility. Considering the latter reason, is the color pattern of the lunar surface not similar to a rabbit? Here, the similarity between rabbit and the lunar surface pattern was evaluated using seven AI architectures. In the test conducted with \textbf{Contrastive Language-Image Pre-Training (CLIP), which can classify images based on given words,} it was assumed that people frequently observe the Moon in the early evening. Under this condition, the lunar surface pattern was found to be more similar to a rabbit than a face in low-latitude regions, while it could also be classified as a face as the latitude increases. This result is consistent with that the oldest literatures about the Moon rabbit were written in India and that a tradition of seeing a human face in the Moon exists in Europe. In a 1000-class test using seven AI architectures, ConvNeXt and CLIP sometimes classified the lunar surface pattern as a rabbit with relatively high probabilities. Cultures are generated by our attitude to the environment. Both dynamic and static similarities may be essential to induce our imagination.
\\
\\
\textbf{Keywords:} Artificial intelligence $\cdot$ ~Lunar surface $\cdot$ ~Rabbit $\cdot $~Face $\cdot$ ~Cognition
\\
\newpage
\section{Introduction}
The Moon is one of the main subjects in human culture, along with the Sun. When we look at the Moon, we can recognize that the surface of the Moon is differentiated into two regions, bright and dark areas. This difference is caused by the different composition of lunar rock. The bright region, known as the highland, is composed of anorthositic rock (e.g., Taylor et al. 1991). In contrast, the dark region, named the mare, is composed of basaltic rock, which was generated inside the Moon and covered the lunar surface through volcanic eruptions (e.g., Taylor et al. 1991). This contrast in surface pattern has fascinated people throughout history and has been regarded as several objects in different cultures. For example, in European countries, there is a tradition that the color pattern of the lunar surface is human's face. In the 1st century, Plutarch discussed the face in the Moon (Pasachoff et al. 2009). During the Renaissance, Giotto di Bondone painted the Moon with a face in his work \textit{the Last Judgment} (Pasachoff et al. 2009). 

\begin{figure}[h]
\centering
\includegraphics[width=14cm] {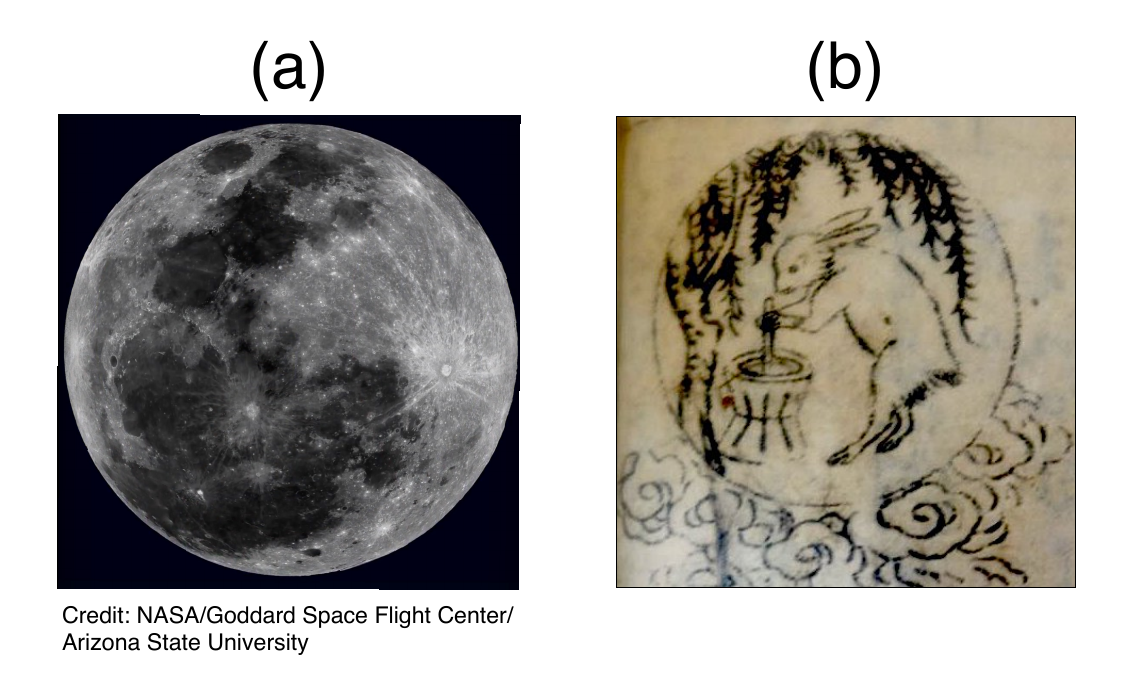}
\caption{(a): Contrast of the surface of the Moon. (b): Illustration of the Moon rabbit in the Japanese book \textit{Zouho-houryaku-ohzassho} (1781 CE, Author’s collection). While the Moon rabbit makes an elixir of immortality in Chinese culture, in Japanese tradition, the Moon rabbit is pounding rice cake on the Moon. On the left side of the rabbit, there is an osmanthus fragrans tree, which is also dipicted as an object on the Moon in Asian tradition.}
\label{picture}
\end{figure}

While the lunar surface has been associated with a face in Europe, in Asian countries such as China and Japan, the lunar surface has been interpreted as the shape of a rabbit/hare (Fig. \ref{picture}). Recently, China launched rovers to the Moon. These rovers were named \textit{Yu-tu} (Jade Rabbit), after the name of the Moon rabbit (Lakdawalla, 2014). The tradition of the Moon rabbit is very old. We can see the sentences that mention the Moon rabbit in the Indian manuscripts  \textit{Jainimiya Brahmana} and \textit{Satapatha Brahmana}, which are thought to have been edited more than 2500 years ago (Bodewitz, 1973; Eggeling, 1900). China has also had the culture of the Moon rabbit for more than 2,000 years. From the Mawangdui tomb, constructed in the 2nd BCE during the Han dynasty, T-shaped silks dipicting the Moon rabbit have been discovered (Kelley, 1978). 

\begin{figure}[h]
\centering
\includegraphics[width=15cm] {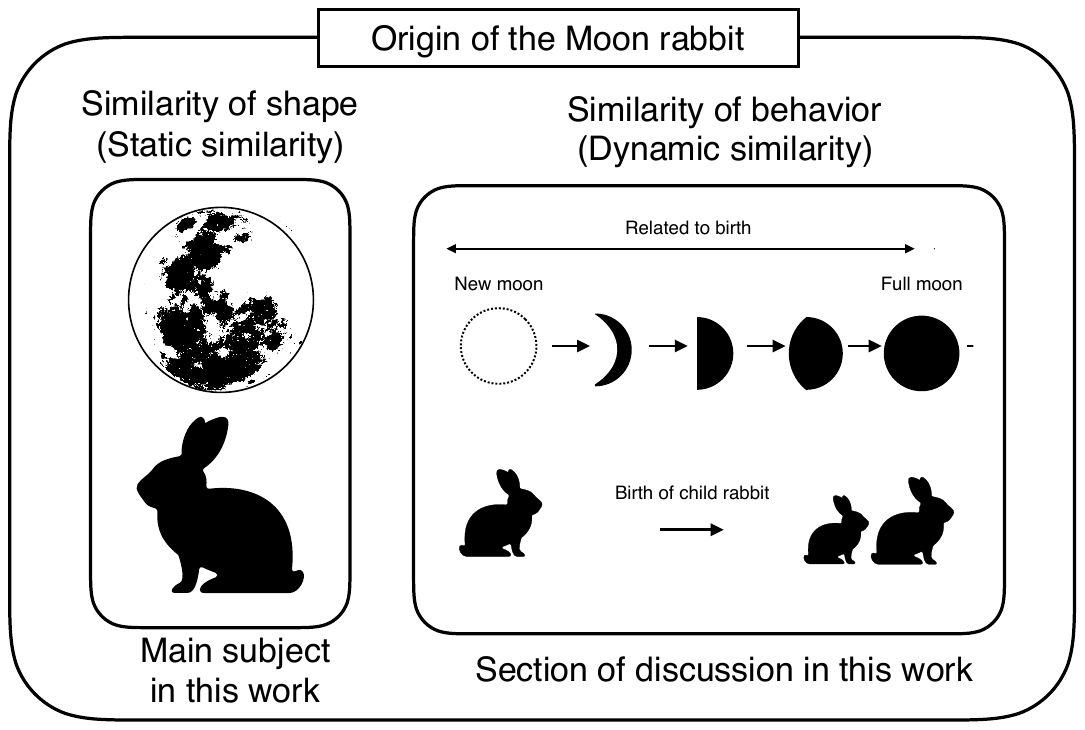}
\caption{Two types of similarity between the rabbit and the Moon as the origin of the Moon rabbit. It has been suggested that the Moon rabbit originated from the pattern of the lunar maria, which is similar to the shape of a rabbit. In cultural anthropology, it has been indicated that the origin of the Moon rabbit is also linked to the cyclic appearance of rabbits and the Moon. While the similarity of shapes is considered static similarity, the relationship of their behaviors can be termed dynamic similarity. In this work, the main focus is on static similarity, as evaluated by the AI architectures. The effects of dynamic similarity are discussed in the discussion section.}
\label{origin}
\end{figure}

It is commonly stated that the culture of the Moon rabbit arose from the similarity between the pattern of the lunar surface and a rabbit's shape. In psychology, the phenomenon in which we tend to perceive a face or object in an ambiguous pattern is called \textit{pareidolia} (e.g., Palmisano et al. 2023; Zhou and Meng, 2020). However, this may not be the sole explanation. In cultural anthropology, it has been suggested that the Moon rabbit was caused by the behavior of these two entities rather than by their visual similarity (e.g., Thuillard and Le Quellec, 2017; Thuillard, 2021). A characteristic behavior of the Moon is that it waxes and wanes (i.e., the Moon appears and disappears cyclically). This cyclic change of the Moon can be associated with the concept of "birth" and "death." The rabbit, known for bearing many offspring, also exhibits a form of cyclic behavior. Cultural anthropologists have hypothesized that this behavioral similarity connected the Moon and the rabbit as common symbols of fertility. In Chinese tradition, for instance, the Moon rabbit is said to make an elixir of immortality. While the shape relationship is a "static similarity," the similarity in behavior can be described as a "dynamic similarity" (Fig. \ref{origin}).

If we consider the hypothesis that the Moon rabbit is based on behavioral similarity, a question arises as to how closely the color pattern of the lunar surface resembles a rabbit. Is the similarity of shape not a crucial factor in the formation of the Moon rabbit culture? This question also related to the ways in which our cultures are formed. To address this question, we must examine the similarity between a rabbit and the lunar surface. However, people have biases influenced by their cultural backgrounds, making unbiased shape comparisons a challenging task.

To analyze obscure similarities, AI technologies may be helpful. Recent AI architectures with supervised training have already been able to classify objects with the same accuracy as humans. In addition, the architecture named CLIP (Contrastive Language-Image Pre-Training), developed by OpenAI, can classify objects into untrained classes (Radford et al. 2021). These AI architectures can output probabilities indicating the likelihood that input objects belong to each class, which is useful for quantitatively evaluating similarities between objects. Shoji (2017) conducted a classification of the lunar surface using a convolutional neural network, considering three classes: "rabbit," "lion," and "crab." However, Shoji (2017) employed a small neural network trained on only a few hundred images. Given that recent architectures are trained on datasets containing tens of millions of images, classification by AI should be updated. 

In this work, using state-of-the-art AI architectures (CLIP, ResNet-50, ViT Large Patch 16, BiT ResNetV2 152$\times$2, SWSL ResNeXt-101 32$\times$8d, ConvNeXt Large, and Noisy Student), the similarity between the lunar surface pattern and a rabbit is evaluated (Radford et al. 2021; He et al. 2016; Dosovitskiy et al. 2020; Kolesnikov et al. 2020; Yalniz et al. 2019; Liu et al. 2022; Xie et al. 2020). For this evaluation, two types of tests were conducted. The first is the rabbit-face test with CLIP. In this test, silhouette images of lunar maria in different orientations were input to CLIP, and the probabilities that they were classified into the two classes, "rabbit" and "face," were calculated. This test was conducted to evaluate the relationship between latitude (where people observe the Moon) and lunar-related cultures. Thuillard (2021) reports the regions where the cultures of the Moon rabbit and the human face in the Moon exist. While the Moon rabbit tradition is predominantly found at low latitudes, such as in southern Asia and Mexico, many regions with a tradition of seeing a human face in the Moon are located at relatively high latitudes, although some low-latitude areas also perceive a human face in the Moon (and there are high-latitude regions that envision the Moon rabbit as well).
How the full Moon appears to us depends on time, season, and latitude because the relative angle of observation changes with these factors. Fig. \ref{posture} shows the appearance of the full Moon in 500 BCE at different times and latitudes, simulated by Stellarium (\url{https://stellarium.org}). Assuming that people frequently view the Moon in the early evening rather than at other times, such as midnight, we hypothesize that the lunar surface around approximately 8:00 PM may resemble a rabbit at low latitudes, while a human face may be perceived at high latitudes. This hypothesis is tested using CLIP.

The second test was conducted with models pre-trained on ImageNet. ImageNet is one of the most widely used datasets for AI classification (Deng et al. 2009), and weights trained on ImageNet are available for several AI architectures. In this study, the seven architectures (CLIP, ResNet-50, ViT Large Patch 16, BiT ResNetV2 152$\times$2, SWSL ResNeXt-101 32$\times$8d, ConvNeXt Large, and Noisy Student) were used to determine the classes into which the lunar surface is categorized. Based on the results of these two tests, we discuss the origin of the Moon rabbit (Fig. \ref{origin}).

\section{Method}
\subsection{Preparation for lunar images}
\begin{figure}[h]
\centering
\includegraphics[width=15cm] {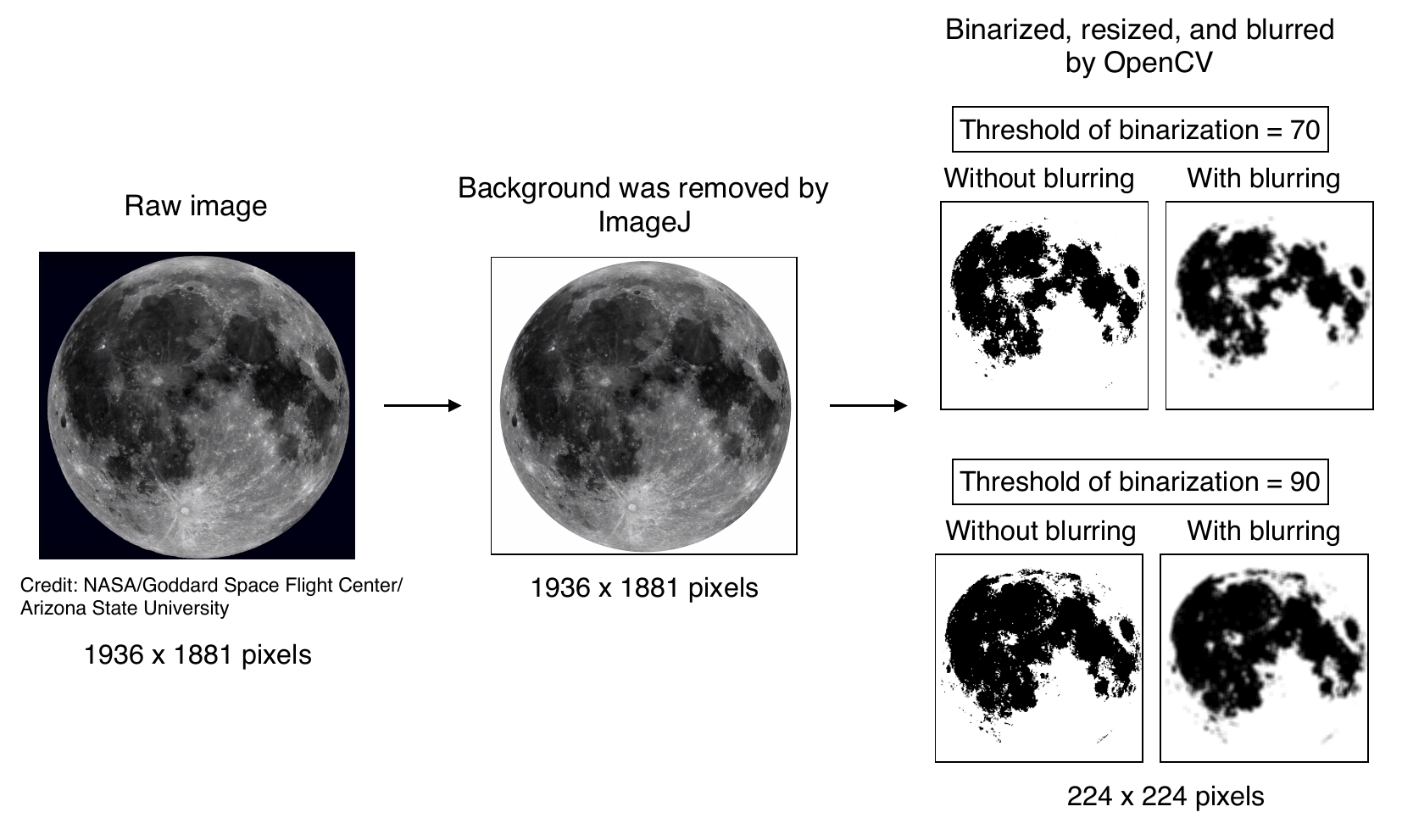}
\caption{Process of image preparation. Using OpenCV, four types of images were prepared.}
\label{image}
\end{figure}

In this work, a lunar image from NASA Solar System Exploration was used as the raw image (\url{https://solarsystem.nasa.gov/system/downloadable_items/3094_lro_nearside.jpg}). This image has dimensions of 1936$\times$1881 pixels. The black background was removed using ImageJ (Schneider et al. 2012) (Fig. \ref{image}). The AI architectures considered in this work are configured to accept input images of 224$\times$224 pixels. To preserve the aspect ratio of the Moon, the image was cropped from its center to 1881$\times$1881 pixels using OpenCV (Bradski, 2000), removing side margins that do not contain the Moon. This cropping did not alter the Moon's shape and resulted in an image showing the circular shape of the Moon itself.
When this image was input to the AI architectures, classification results were strongly affected by its circular shape (i.e., the AI tended to classify the Moon as a circular object). The purpose of this work, however, is to classify the pattern of the lunar maria rather than the circular edge of the Moon. To isolate the lunar maria pattern, the image was converted to grayscale and then binarized, leaving only the maria areas in black. Binarization thresholding is a crucial factor as it determines the area extracted as lunar maria. Here, two threshold values, 70 and 90, were tested. Pixels with values below these thresholds were changed to 0 (black), while others were set to 255 (white). The binarized images are shown in Fig. \ref{image}. For threshold = 90, a larger area was extracted as lunar maria compared with a threshold of 70. It is important to note that the binarized images have a uniform pixel value (0) inside the maria regions. However, contrast within the maria is not essential for this work, as the study focuses on the pattern rather than the texture of the maria. Finally, the binarized images were resized to 224$\times$224 pixels.

These resized images served as the base images for all tests. In addition to these images, blurred images were prepared using a 5$\times$5 kernel filter (Fig. \ref{image}). The blurred images emphasize the global distribution of the maria, as this global patterning is typically principal information for humans imagining an object. In the blurred images, local details of each mare were smoothed, highlighting the global patterns compared with the base images. In each test, assuming that people observe the Moon most frequently in the early evening, the images were rotated to match the Moon’s orientation at 8:00 PM at various latitudes in January (Fig. \ref{posture}). Although the Moon’s orientation changes seasonally, the January posture was chosen because the pattern during this season (winter and spring) most closely resembles an upright rabbit shape at low latitudes, where the lunar pattern is most likely interpreted as a rabbit (Fig. \ref{picture}). Every lunar image used in this work is stored at \url{https://github.com/Enceladus47/Classification_of_lunar_surface/tree/main}.

\subsection{AI architectures}
In this work, all tests were performed with Google Colaboratory, where AI architectures and pre-trained weights were downloaded. All code used in this work is stored at \url{https://github.com/Enceladus47/Classification_of_lunar_surface/tree/main}. CLIP (Radford et al. 2021), which was used for both the rabbit-face test and the ImageNet test, was downloaded from GitHub ({\url{https://github.com/openai/CLIP}}) to Google Colaboratory. CLIP supports several types of architectures for classification. Here, the pre-trained weights of ViT-B/32 were used. In the rabbit-face test with CLIP, by providing the words “rabbit” and “face,” probabilities for the two classes were calculated.

For the ImageNet test, seven AI architectures -- CLIP ViT-B/32, Resnet 50, ViT Large Patch 16, Bit ResnetV2 152$\times$2, SWSL ResNeXt-101 32$\times$8d, ConvNeXt Large, and Noisy Student -- were selected. Resnet 50 is one of the most commonly used and tested architectures for classification (He et al. 2016). The other six architectures have also been tested for classification in many studies and have shown strong performance in classifying even obscure images, such as silhouettes, compared with other architectures (Geirhos et al. 2021). As mentioned above, this work tested lunar surface patterns using silhouette images, making these architectures suitable choices for the ImageNet test. The six architectures, other than CLIP, were implemented using timm (Wightman, 2019). Table \ref{weight} shows the names of each model used with timm. The purpose of this work is not to classify objects into their correct classes but to identify classes with patterns similar to that of the lunar surface. Therefore, probabilities were considered up to the 10th highest class among the 1000 categories. Since CLIP does not have fixed classes, its probabilities were calculated by assigning the same classes as ImageNet 1K as text words.

\section{Result}
\subsection{Rabbit-face test}
\begin{figure}[h]
\centering
\includegraphics[width=16cm] {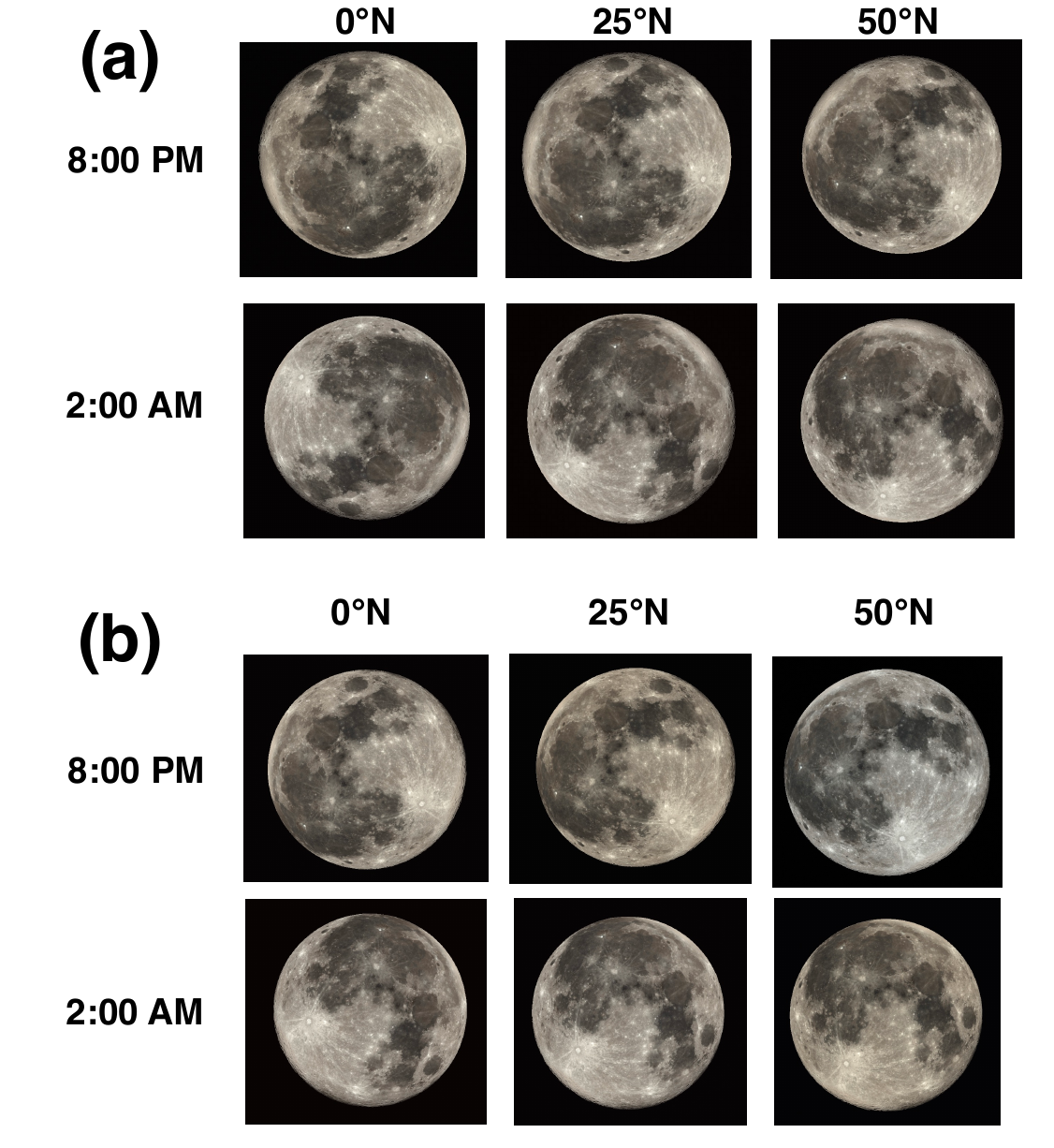}
\caption{Postures of the full Moon simulated by Stellarium (https://stellarium.org) at different times and latitudes in January (a) and July (b) in 500 BCE. The times indicate the local time at each latitude. The images of the Moon are cited from Stellarium under the public domain licenses of NASA \& JPL. }
\label{posture}
\end{figure}

\begin{figure}[h]
\centering
\includegraphics[width=18cm] {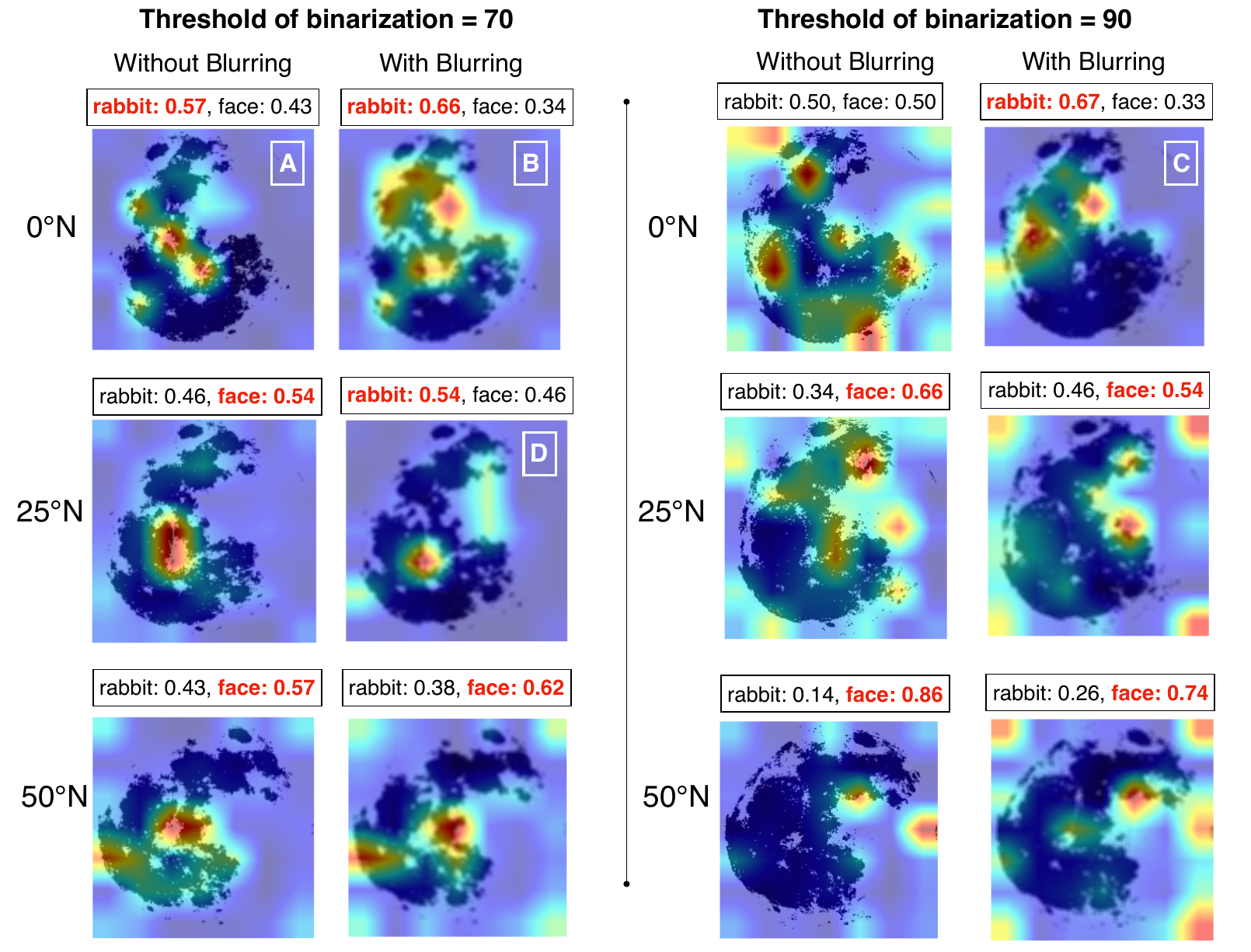}
\caption{Tested lunar images and probabilities of "rabbit" and "face" judged by CLIP. Selected classes (those with higher probabilities) are shown in bold red. Attention maps for the selected class are overlaid on each image. The red areas indicate the regions that CLIP focused on when selecting the classes. At a threshold of 90 without blurring, an attention map for "face" is shown because the probabilities for both classes are the same. "A," "B," "C," and "D" in the images selected as "rabbit" correspond to the images tested in the ImageNet test, as shown in Table \ref{imagenet}.}
 \label{rabbit_face}
\end{figure}

Figure \ref{rabbit_face} shows the probabilities of the lunar images for the two classes “rabbit” and “face.” The postures of the images correspond to the Moon at 8:00 PM in 500 BCE at different latitudes (Fig. \ref{posture}). Each image is overlaid with attention maps indicating the most attended areas, generated using the code by Chefer et al. (2021) (\url{https://github.com/hila-chefer/Transformer-MM-Explainability/blob/main/CLIP_explainability.ipynb}). At both thresholds, the probabilities for the “face” class increase with increasing latitude. With a threshold of 70, the lunar image is classified as “rabbit” at 0°N latitude. With a threshold of 90 and without blurring, the probabilities for both classes were the same at 0°N latitude. Thus, CLIP tends to classify the lunar surface pattern as similar to a rabbit rather than a face at low latitudes, while the pattern is relatively regarded as a face at high latitudes. From the attention maps, at low latitudes, CLIP focuses on the regions associated with the rabbit’s face and body (Fig. \ref{picture}).
In contrast, CLIP is attracted to the central area or periphery of the pattern at high latitudes. As latitude increases, the rabbit's shape lies down throughout the seasons (Fig. \ref{posture}), making it difficult even for humans to perceive a rabbit. Thus, CLIP may also classify the lunar images as a face rather than a rabbit. Several human tests also show that the response to a face weakens when face-like images are displayed in an inverted orientation (Romagnano et al. 2023).

Comparing the images with and without blurring, the probabilities for “rabbit” increase at 0°N and 25°N latitude (Fig. \ref{posture}). When localized changes are reduced, the lunar surface pattern is more clearly perceived as a rabbit. However, at 50°N latitude, the images are consistently classified as “face” with a probability greater than 0.6, regardless of blurring. At this posture, the global shape of the pattern may make it difficult to perceive a rabbit.

Assuming that people view the Moon more frequently in the early evening rather than at midnight or early morning, the results of the rabbit-face test may offer insight into why the Moon rabbit culture exists in Asian countries, while European culture tends to see a face in the Moon. As CLIP’s classification indicates, the probability of “rabbit” decreases as the area associated with the rabbit’s face lies down. For ancient people living at high latitudes, it may have been difficult to imagine a rabbit shape from the lunar pattern in the early evening. Conversely, for those at lower latitudes, the lunar pattern may readily suggest the shape of a rabbit. As mentioned in the introduction, the oldest texts mentioning the Moon rabbit are found in the Indian literatures. The oldest Chinese text that references the Moon rabbit, \textit{Chu ci}, was also written in southern China, \textit{Chu}. The fact that the earliest literatures about the Moon rabbit originate from areas at relatively lower latitudes is consistent with the results of the CLIP test.

In cultural anthropology, the origin of the Moon rabbit is typically associated with its symbolism. However, in addition to the behavior (dynamic similarity) of the Moon and rabbit, CLIP’s classifications suggest that the similarity in shape (static similarity) between objects may also contribute to the formation of cultural symbols.

\subsection{ImageNet test}
In a subsequent test, images classified as “rabbit” in the rabbit-face test were evaluated by selecting the top 10 classes of ImageNet 1K using seven architectures. In addition to the top 10 classes, the probability of the “rabbit” class was also calculated. ImageNet includes three rabbit-related classes: “Angora,” “hare,” and “wood\_rabbit.” Since this study does not focus on rabbit species classification, the overall probability of the rabbit class was calculated as the sum of these three classes. Table \ref{imagenet} shows that the selected classes varied across architectures. Several classes (e.g., Nematode, Tick, and Petri dish) were frequently selected with high probability by all architectures. These classes likely appeared due to their visual similarity to the silhouetted lunar images. Interestingly, CLIP selected the “tailed frog” class within its top 10 choices, which is notable since frog/toad is also an animal traditionally associated with the Moon in Chinese folklore.

Overall, the probability of the “rabbit” class was much lower than that of the top 10 selected classes. This indicates that the AI architectures did not consistently interpret the lunar images as resembling a rabbit. However, there were some exceptions. In the case of image C, CLIP assigned a “rabbit” probability of approximately 0.02, making it the second-most probable class (Table \ref{imagenet}). Here, the “Angora” class probability was notably high ($\sim$0.018), contributing to the overall probability for the rabbit class. This image features a large area of lunar maria with a blurred pattern, potentially resembling the texture of Angora fur. Additionally, CLIP assigned a relatively high probability to the “rabbit” class in image B, placing it among the seventh-most probable classes (Table \ref{imagenet}). ConvNeXt also assessed the probability of “rabbit” in image D, finding it comparable to the tenth class, “Ping-pong ball.”
Given the 1000 classes in ImageNet 1K, these results suggest that, although “rabbit” was not typically the most probable class, the AI sometimes judged the lunar surface as not entirely dissimilar from a rabbit’s shape. For imagination to be sparked, relatively probable classes (not necessarily the top class) play an essential role, as the top-ranked class may be too similar to invite associations with other objects. The highest-probability class is typically seen as the “same” object rather than a “similar” one. The selection of the rabbit class with a relatively high probability implies that the lunar pattern may indeed have the potential to evoke a rabbit-like image. Unlike AI, humans can communicate through language. Once someone tells others that the surface of the Moon resembles a rabbit, this suggestion may influence others to begin seeing a rabbit. In this way, humans can adapt their “architectures” or mental frameworks through communication.

Of course, the dynamic pattern (behavior) cannot be ruled out when a particular object is linked to the Moon. As cultural anthropologists have indicated, ancient people were likely attracted to the cyclic changes of the Moon and the fertility of rabbits. The relationship between static and dynamic (shape and behavior) similarities is discussed below.

\section{Discussion}
\subsection{Effect of dynamic similarity}
This study primarily focused on shape similarity (i.e., static similarity). The AI tests conducted in this study indicate that the lunar surface pattern could resemble a rabbit, especially for observers in low-latitude regions. However, the probability of classifying the pattern as a rabbit was not so high. The ImageNet test shows that many classes were more probable than the rabbit class. Cultural anthropologists suggest that the cyclical waxing and waning of the Moon and the rabbit's association with fertility link them symbolically. Consequently, we cannot exclude discussions about the role of dynamic similarity, as emphasized by cultural anthropologists.

French philosopher Gilles Deleuze says, ” There are greater differences between a plow horse or draft horse and a racehorse than between an ox and a plow horse” (Deleuze, 1988). A plow horse is a horse for agriculture. An ox is also used for agricultural work. Thus, when we consider the role (function) of animals, a plow horse is classified into the same classes as an ox, rather than a racehorse. In this classification, the criterion is not the shape of the horse and ox but the types of work, "plow" and "race." In cognitive semantics, foundational relationships and movements that shape human cognition have been explored. For instance, Johnson (1987) introduces several cognitive patterns (image schemas) such as "path" and "containment." Cyclic pattern, like recurrence, is also part of these schemas. Aristotle, the ancient Greek philosopher, describes two primary types of movement -- linear and circular -- in his book \textit{On the Heavens}. The defining feature of circular movement is its recurrence, returning to its starting point. As cyclic movement is foundational for human cognition, the Moon's cycle (waxing and waning) may have naturally connected it to other cyclically perceived entities, such as the rabbit, known for its fertility. If ancient observers associated the Moon's cyclic changes with the fertility of rabbits, the two objects could have been linked through the cognitive schema of "cycle." Once this conceptual link was formed, it may have influenced people to see shape similarities between the Moon and a rabbit. However, as the AI tests suggested, at higher latitudes, it is more challenging to see a rabbit shape in the Moon’s appearance, even with a “cycle” schema in mind. Consequently, the Moon rabbit association might have first emerged in lower-latitude regions where the Moon’s orientation more readily resembles a rabbit.

Once established, however, cultures can spread. Cultural transmission is influenced by several factors beyond fundamental cognition, such as political relationships between regions. In the case of the Moon rabbit, the propagation of Buddhism significantly contributed to its widespread adoption, as Buddhist literature, the \textit{Jataka} tales, contains a story about the Moon rabbit. Thus, the culture of the Moon rabbit could also spread to northern regions.

\subsection{Classification of dynamic patterns by AI}
If dynamic similarities play an important role, to understand the culture of the Moon rabbit more comprehensively, we must consider classifications based on dynamic patterns. Along with image classifications, machine learning techniques for classifying video frames have made significant progress, particularly in detecting the types of actions of each object (e.g., Karpathy et al., 2014). Although there are many challenges yet to be solved, such as the different time scales of patterns (the birth of rabbits and the waxing of the Moon occur on different time scales), if we can apply these techniques and video technologies, such as time-lapse, we may be able to attempt AI classifications that consider both static and dynamic patterns (Shoji et al., 2022). Combining AI technologies with discussions in cultural anthropology will be an important topic for understanding our cultures.

\section{Conclusion}
Recent AI architectures have made significant progress. In the case of image classification, machines can classify objects with accuracy comparable to that of humans. In this study, the shape similarity between the rabbit and the Moon was evaluated using two types of tests. Using the two classes, "rabbit" and "face," the probability of "rabbit" tended to increase with decreasing latitude, where people observe the Moon. This result is consistent with the fact that the oldest texts mentioning the Moon rabbit were written in India and southern China. Considering the 1000 classes of ImageNet 1K, the probability of "rabbit" was not high. However, both ConvNeXt and CLIP sometimes concluded that the pattern of the lunar surface was not so different from the shape of a rabbit. Thus, some ancient people at low latitudes might also have imagined a rabbit's body when they looked at the Moon. However, as anthropologists have indicated, the dynamic pattern of the Moon and the rabbit must also played an important role. Therefore, further studies are needed to understand the generation of culture. Applying video classification techniques may be helpful in considering the dynamic patterns of objects.

Although it is beyond the scope of this work, there is another factor that induces culture: wonder about the environment. As philosophers say, wonder is the foundation of our cognition and curiosity. Ancient people must have wondered about the behaviors of rabbits and the Moon. Thus, the discussions in this work are still superficial. To induce the feeling of wonder, we must experience ordinary life. We can wonder when we notice events that differ from our everyday experiences. Ancient people found the behaviors of the Moon and the rabbit interesting because they differed from people's daily lives. Culture is a product of our attitude toward the environment (nature and community). I do not know whether machines will wonder in the future. However, studies with AI technologies will provide deeper insights into how we interact with our environment.
\\
\\
\textbf{Funding} No funding, grants, or other support were received.
\\
\\
\textbf{Data availability}  Every lunar image used in this work is deposited in \url{https://github.com/Enceladus47/Classification_of_lunar_surface/tree/main}.

\section*{Declarations}
\textbf{Conflict of interest} The author has no competing interests to declare that are relevant to the content of this article.

\nocite{*}

\begin{thebibliography}{29}
\providecommand{\natexlab}[1]{#1}
\providecommand{\url}[1]{{#1}}
\providecommand{\urlprefix}{URL }
\expandafter\ifx\csname urlstyle\endcsname\relax
  \providecommand{\doi}[1]{DOI~\discretionary{}{}{}#1}\else
  \providecommand{\doi}{DOI~\discretionary{}{}{}\begingroup
  \urlstyle{rm}\Url}\fi
\providecommand{\eprint}[2][]{\url{#2}}

\bibitem[{Bodewitz(1973)}]{bodewitz19731}
Bodewitz HW (1973) Jaiminiya Brahmana I, 1-65. Leiden: Brill

\bibitem[{Bradski(2000)}]{opencv_library}
Bradski G (2000) {The OpenCV Library}. Dr Dobb's Journal of Software Tools

\bibitem[{Chefer et~al(2021)Chefer, Gur, and Wolf}]{Chefer_2021_ICCV}
Chefer H, Gur S, Wolf L (2021) Generic attention-model explainability for
  interpreting bi-modal and encoder-decoder transformers. In: Proceedings of
  the IEEE/CVF International Conference on Computer Vision (ICCV), pp 397--406

\bibitem[{Deleuze(1988)}]{deleuze1988spinoza}
Deleuze G (1988) Spinoza: practical philosophy. City Lights Books

\bibitem[{Deng(2009)}]{deng2009large}
Deng J (2009) A large-scale hierarchical image database. Proc of IEEE Computer
  Vision and Pattern Recognition, 2009

\bibitem[{Dosovitskiy et~al(2020)Dosovitskiy, Beyer, Kolesnikov, Weissenborn,
  Zhai, Unterthiner, Dehghani, Minderer, Heigold, Gelly
  et~al}]{dosovitskiy2020image}
Dosovitskiy A, Beyer L, Kolesnikov A, Weissenborn D, Zhai X, Unterthiner T,
  Dehghani M, Minderer M, Heigold G, Gelly S, et~al (2020) An image is worth
  16x16 words: Transformers for image recognition at scale. arXiv preprint
  arXiv:201011929

\bibitem[{Eggeling(1900)}]{eggeling2014satapatha}
Eggeling J (1900) The Satapatha Br{\^a}hmana, part V, vol~1. The Clarendon
  Press

\bibitem[{Geirhos et~al(2021)Geirhos, Narayanappa, Mitzkus, Thieringer, Bethge,
  Wichmann, and Brendel}]{geirhos2021partial}
Geirhos R, Narayanappa K, Mitzkus B, Thieringer T, Bethge M, Wichmann FA,
  Brendel W (2021) Partial success in closing the gap between human and machine
  vision. Advances in Neural Information Processing Systems 34:23,885--23,899

\bibitem[{He et~al(2016)He, Zhang, Ren, and Sun}]{he2016deep}
He K, Zhang X, Ren S, Sun J (2016) Deep residual learning for image
  recognition. In: Proceedings of the IEEE conference on computer vision and
  pattern recognition, pp 770--778

\bibitem[{Johnson(1987)}]{Johnson1987}
Johnson M (1987) The Body in the Mind: The Bodily Basis of Meaning,
  Imagination, and Reason. University of Chicago press

\bibitem[{Karpathy et~al(2014)Karpathy, Toderici, Shetty, Leung, Sukthankar,
  and Fei-Fei}]{karpathy2014large}
Karpathy A, Toderici G, Shetty S, Leung T, Sukthankar R, Fei-Fei L (2014)
  Large-scale video classification with convolutional neural networks. In:
  Proceedings of the IEEE conference on Computer Vision and Pattern
  Recognition, pp 1725--1732

\bibitem[{Kelley(1978)}]{kelley1978examination}
Kelley CW (1978) An examination of the t-shaped painting from the western han
  tomb no. 1 at ma-wang-tui, ch'ang-sha, hunan. PhD thesis, University of
  British Columbia

\bibitem[{Kolesnikov et~al(2020)Kolesnikov, Beyer, Zhai, Puigcerver, Yung,
  Gelly, and Houlsby}]{kolesnikov2020big}
Kolesnikov A, Beyer L, Zhai X, Puigcerver J, Yung J, Gelly S, Houlsby N (2020)
  Big transfer (bit): General visual representation learning. In: Computer
  Vision--ECCV 2020: 16th European Conference, Glasgow, UK, August 23--28,
  2020, Proceedings, Part V 16, Springer, pp 491--507

\bibitem[{Lakdawalla(2014)}]{lakdawalla2014china}
Lakdawalla E (2014) China lands on the moon. Nature Geoscience 7(2):81--81

\bibitem[{Liu et~al(2022)Liu, Mao, Wu, Feichtenhofer, Darrell, and
  Xie}]{liu2022convnet}
Liu Z, Mao H, Wu CY, Feichtenhofer C, Darrell T, Xie S (2022) A convnet for the
  2020s. In: Proceedings of the IEEE/CVF conference on computer vision and
  pattern recognition, pp 11,976--11,986

\bibitem[{Palmisano and et~al.(2023)}]{palmisano2023}
Palmisano A, et~al (2023) Face pareidolia is enhanced by 40 hz transcranial
  alternating current stimulation (tacs) of the face perception network.
  Scientific Reports 13(1):2035

\bibitem[{Pasachoff and Olson(2019)}]{pasachoff2019depictions}
Pasachoff JM, Olson RJ (2019) Depictions of the moon in western visual culture.
  In: Oxford Research Encyclopedia of Planetary Science

\bibitem[{Radford et~al(2021)Radford, Kim, Hallacy, Ramesh, Goh, Agarwal,
  Sastry, Askell, Mishkin, Clark et~al}]{radford2021learning}
Radford A, Kim JW, Hallacy C, Ramesh A, Goh G, Agarwal S, Sastry G, Askell A,
  Mishkin P, Clark J, et~al (2021) Learning transferable visual models from
  natural language supervision. In: International conference on machine
  learning, PMLR, pp 8748--8763

\bibitem[{Romagnano et~al(2023)Romagnano, Sokolov, Fallgatter, and
  Pavlova}]{romagnano2023subtle}
Romagnano V, Sokolov AN, Fallgatter AJ, Pavlova MA (2023) Do subtle cultural
  differences sculpt face pareidolia? Schizophrenia 9(1):28

\bibitem[{Schneider et~al(2012)Schneider, Rasband, and
  Eliceiri}]{schneider2012nih}
Schneider CA, Rasband WS, Eliceiri KW (2012) Nih image to imagej: 25 years of
  image analysis. Nature methods 9(7):671--675

\bibitem[{Shoji(2017)}]{shoji2017does}
Shoji D (2017) What does a convolutional neural network recognize in the moon?
  arXiv:170805636

\bibitem[{Shoji(2022)}]{shoji2022rabbit}
Shoji D (2022) Rabbit, toad, and the moon: Can machine categorize them into one
  class? arXiv preprint arXiv:220316163

\bibitem[{Taylor et~al(1991)Taylor, Warren, Ryder, Delano, Pieters, and
  Lofgren}]{taylor1991lunar}
Taylor GJ, Warren P, Ryder G, Delano J, Pieters C, Lofgren G (1991) Lunar
  rocks. Lunar Sourcebook, A User's Guide to the Moon pp 183--284

\bibitem[{Thuillard(2021)}]{thuillard2021analysis}
Thuillard M (2021) Analysis of the worldwide distribution of the ‘man or
  animal in the moon’motifs. Folklore: Electronic Journal of Folklore
  (84):127--144

\bibitem[{Thuillard and Le~Quellec(2017)}]{thuillard2017phylogenetic}
Thuillard M, Le~Quellec JL (2017) A phylogenetic interpretation of the
  canonical formula of myths by l{\'e}vi-strauss. Cultural Anthropology and
  Ethnosemiotics 3(2):1--12

\bibitem[{Wightman(2019)}]{rw2019timm}
Wightman R (2019) Pytorch image models.
  \url{https://github.com/rwightman/pytorch-image-models},
  \doi{10.5281/zenodo.4414861}

\bibitem[{Xie et~al(2020)Xie, Luong, Hovy, and Le}]{xie2020self}
Xie Q, Luong MT, Hovy E, Le QV (2020) Self-training with noisy student improves
  imagenet classification. In: Proceedings of the IEEE/CVF conference on
  computer vision and pattern recognition, pp 10,687--10,698

\bibitem[{Yalniz et~al(2019)Yalniz, J{\'e}gou, Chen, Paluri, and
  Mahajan}]{yalniz2019billion}
Yalniz IZ, J{\'e}gou H, Chen K, Paluri M, Mahajan D (2019) Billion-scale
  semi-supervised learning for image classification. arXiv preprint
  arXiv:190500546

\bibitem[{Zhou and Meng(2020)}]{zhou2020you}
Zhou LF, Meng M (2020) Do you see the “face”? individual differences in face
  pareidolia. Journal of Pacific Rim Psychology 14:e2

\end{thebibliography}

\newpage

\begin{table}[h]
\begin{center}
\caption{Name of models used with timm.}
\label{weight}
\begin{tabular}{l l}
\hline 
Architecture & Name used with timm\\ \hline
Resnet 50 & resnet50\\
ViT Large Patch 16 & vit\_large\_patch16\_224\\
Bit ResnetV2 152$\times$2 & resnetv2\_152x2\_bitm\\
SWSL ResNeXt-101 32$\times$8d &swsl\_resnext101\_32x8d\\
ConvNeXt Large & convnext\_large\_in22ft1k\\
Noisy Student & tf\_efficientnet\_b0\_ns \\
\hline
\end{tabular}
\end{center}
\end{table}

\newpage
\begin{table}[h]
\caption{Top 10 classes and their probabilities for each image selected by the seven AI architectures. The probabilities of the rabbit class are also shown at the end of each table. Rabbit classes with probabilities comparable to the top 10 classes are highlighted with blue backgrounds.}
\label{imagenet}
\centering
\includegraphics[width=15cm] {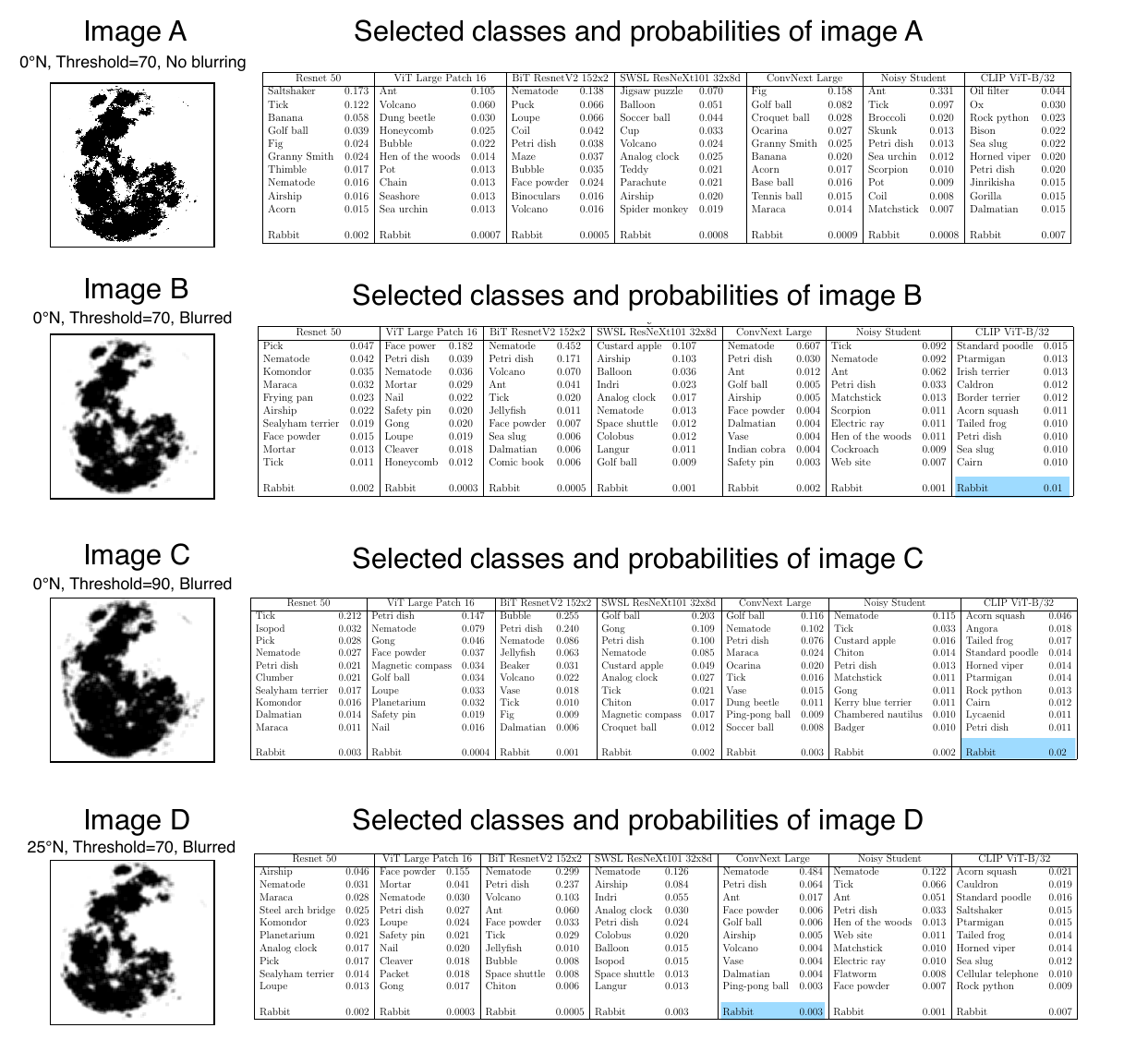}
\end{table}
\newpage

\newpage

\end{document}